# LPYOLO: Low Precision YOLO for Face Detection on FPGA

**Bestami Gunay[1,2], Sefa Burak Okcu[1], Hasan Sakir Bilge[2]**
[1]Aselsan Inc.
Ankara, Turkey
bgunay@aselsan.com.tr; burakokcu@aselsan.com.tr
[2]Gazi University Electrical Electronics Engineering Department
Ankara, Turkey
bilge@gazi.edu.tr

***Abstract*** - In recent years, number of edge computing devices and artificial intelligence applications on them have advanced excessively. In edge computing, decision making processes and computations are moved from servers to edge devices. Hence, cheap and low power devices are required. FPGAs are very low power, inclined to do parallel operations and deeply suitable devices for running Convolutional Neural Networks (CNN) which are the fundamental unit of an artificial intelligence application. Face detection on surveillance systems is the most expected application on the security market. In this work, TinyYolov3 architecture is redesigned and deployed for face detection. It is a CNN based object detection method and developed for embedded systems. PYNQ-Z2 is selected as a target board which has low-end Xilinx Zynq 7020 System-on-Chip (SoC) on it. Redesigned TinyYolov3 model is defined in numerous bit width precisions with Brevitas library which brings fundamental CNN layers and activations in integer quantized form. Then, the model is trained in a quantized structure with WiderFace dataset. In order to decrease latency and power consumption, on-chip memory of the FPGA is configured as a storage of whole network parameters and the last activation function is modified as rescaled HardTanh instead of Sigmoid. Also, high degree of parallelism is applied to logical resources of the FPGA. The model is converted to an HLS (High-Level-Synthesis) based application with using FINN framework and FINN-HLS library which includes the layer definitions in C++. Later, the model is synthesized and deployed. CPU of the SoC is employed with multithreading mechanism and responsible for preprocessing, postprocessing and TCP/IP streaming operations. Consequently, 2.4 Watt total board power consumption, 18 Frames-Per-Second (FPS) throughput and 0.757 Mean-Average-Precision (mAP) accuracy rate on Easy category of the WiderFace are achieved with 4 bits precision model.

## 1. Introduction

Smart applications on edge devices have increased drastically in last decades with the help of the latest developments on embedded technologies. Face detection on video surveillance systems is one of the most demanded security applications for the intelligent systems. Recently, several methods based on deep neural networks have been advanced for the object detection and these methods can also be deployed for face detection. Object detection algorithms are divided into two categories; two stage object detectors and single stage object detectors. R-CNN (Region Based CNN) [1], Fast R-CNN [2], Faster R-CNN [3] and Mask R-CNN [4] are the two stage object detectors. In the first stage, region proposal is generated and in the second stage, object classification and bounding box regression of each region proposal is done for the given image or video frame. On the other hand, SSD (Single Shot Multibox Detector) [5], YOLO (You Only Look Once) [6], YOLOv2 [7], YOLOv3 [8], YOLOv4 [9], YOLOv5 [10] and CenterNet [11] are the single stage object detectors. In these methods, the model takes the input frame and gives the coordinates of the bounding box and confidence scores of the classes in a single stage. Therefore, single stage detectors are faster, but have generally lower accuracy compared to two stage detectors [12]. Hence, single stage detectors are more convenient for embedded systems which have low processing capabilities.

Deep neural networks might consist of millions of matrix-matrix multiplication and accumulation operations. Therefore, parallel processing capability is required in order to get reasonable throughput rate on CNN deployments. CNN implementations on CPUs give very low speed due to having sequential processing capability with low count cores. As an alternative, GPUs are very fast devices for neural networks, nevertheless they consume too much power [13]. However, cheap and low power devices are needed on edge applications. After all, parallel processing circuits can be designed on the FPGA as a result of having reconfigurable hardware architecture with low power consumption. In this aspect, SoC devices which contain CPU and FPGA in a single chip together are suitable choice for the implementation of deep neural networks because of having small chip size, optimized power consumption and high bandwidth between Processing Subsystem (PS) and Programmable Logic (PL). Moreover, PS of the SoC is good enough for preprocessing and postprocessing operations of a CNN implementation.

This work focuses on inference of modified version of TinyYolov3 for face detection on PYNQ-Z2 board which has a low-end Xilinx Zynq 7020 SoC on it. TinyYolov3 is the lightweight version of Yolov3 and especially developed for embedded systems. In this paper, original TinyYolov3 is redesigned in order to fit into the limited resources of the FPGA. The deep neural network model is created on different integer bit precisions for weights and activations. Then, Quantization Aware Training (QAT) is carried out with Brevitas [14] on WiderFace dataset [15]. The trained model is synthesized for the FPGA with HLS based FINN framework [16]. All the network parameters including weights and activations are stored on FPGA's on-chip memory block RAMs to reduce power consumption and getting higher speed. Then, the model is deployed to the PL side of the SoC with high degree of parallelism applied to the logical resources. The PS side of the SoC is used for preprocessing, postprocessing and TCP/IP streaming operations of the video frames in different threads.

The paper is organized as follows. Section 2 reviews the related works in the literature about the single stage object and face detection algorithms implementation on FPGAs. Proposed architecture and deployment procedure are explained in the Section 3. Later, performance results are given in the Section 4. In the end, the paper is concluded with Section 5.

## 2. Related Works

In literature, if a low-end target device is selected, network parameters are generally stored on an external memory due to having limited resources on the FPGA. Yu and Bouganis [17] targeted a TinyYolov3 model implementation on Zynq 7020 with 16 bits precision. The HLS based CNN architecture runs on the PL. The PS is responsible for the controlling of the system as a baremetal application. Parameters are stored on an external RAM. As a result, 1.88 FPS, 30.9 mAP50 accuracy and 3.36 W power consumption are achieved on COCO dataset. Miranda et al. [18] achieved 30.8 mAP50 accuracy and 14 FPS for 8 bits precision, 31.5 mAP50 accuracy and 7 FPS for 16 bits precision on COCO dataset. Bao et al. [19] also proposed a power efficient Yolov2 architecture with the pipelined network structure. PS runs Ubuntu OS with PYNQ [25] on it. 78.25 mAP accuracy on Pascal VOC dataset and 124 ms (millisecond) latency with 2.7 W power consumption are achieved as test results. Wei et al. [20] tried different method to store network parameters. In order to prevent too much time consumption on external RAM access, on-chip memory BRAM is chosen for storing network parameters. The whole network parameters except input-output layers are binarized. The CNN is constrained with 8 classes of COCO dataset and achieved 0.89 mAP accuracy, 0.4 FPS throughput and 2.8 W power consumption as a result. Huang et al. [21] came up with depthwise deformable convolution on the CenterNet network architecture to get lower resource consumption. ShuffleNetV2 [22] is used as a backbone. 26 FPS throughput, 5.6 W power consumption and 61.7 AP50 accuracy on Pascal VOC dataset are observed. Zynq XCZU3EG is used as target device. Fan et al. [23] targeted implementation of SSDLite with MobileNetV2 [26] backbone on Zynq 7045. Bottleneck residual block technique is used for creating convolution layer. Processing element is shared between other convolutions and it is repeatedly used. Quantization is applied partially, so different bit precisions are used on the different stages of the network. The architecture achieved 20.3 mAP accuracy, 64.8 FPS throughput and 9.9 W power consumption. Umuroglu et al. [24] designed FINN framework to deploy binarized neural networks on FPGA. All the network parameters are used as binary values. That work achieved the fastest neural network inference results on MNIST, CIFAR10 and SVHN datasets. Blott et al. [16] advanced the previous framework FINN to FINN-R. With this framework, TincyYolo model is created and trained in a quantized form with 1 bit for weights and 3 bits for activations. Considering the model is large enough for the Zynq 7020, external RAM is used for storing the network parameters. 2.5 W power consumption and 50.1 mAP accuracy on Pascal VOC dataset are achieved. Xu et al. [27] focused on face detection on a high-end board Xilinx ZCU104. CNN based Single Shot Scale-invariant Face Detector (S³FD) [28] method is employed and, in the results, 37 FPS throughput and 0.918 mAP accuracy on the Easy category of the WiderFace dataset is achieved. Fu et al. [29] deployed MTCNN [30] method for face detection on a high-end board Xilinx ZC706 and achieved 11.9 FPS throughput.

In almost all object or face detection implementations on FPGA, external memory is preferred for the network parameters. Instead, on-chip memory usage can decrease the power consumption and latency. In addition, QAT is used only in [16] and [24] which are FINN framework. A network which is trained with quantized values tends to give better accuracy compared to post-quantized networks. In this paper, we entirely utilized on-chip memory and used FINN framework with QAT.

## 3. Methodology
### 3.1. Low Precision YOLO (LPYOLO) Architecture

Original TinyYolov3 architecture is redesigned as follows. 1. Upsample and concatenation layers are removed. 2. Number of kernels on each convolution layer are five times reduced. 3. QuantConv and QuantReLU, which are in integer quantized form instead of floating point are used as convolution layer and activation function respectively. 4. On the training stage, Sigmoid function is used as activation function of the last layer for better learning. Nevertheless, it is changed to rescaled HardTanh on the synthesis stage for lowering latency and running with FINN. The relationship between Tanh and Sigmoid function can be defined as equation (1).

$$\sigma(x) = \frac{1 + \tanh(x/2)}{2} \quad (1)$$

The CNN model which is shown in Table 1 takes 416x416x3 input image and gives 13x13x18 result which means that there are 13x13 grid for an image and each grid contains center of x and y coordinates, width, height, class information and confidence scores of the detected object. These six outputs are calculated for 3 different anchor boxes which gives the ability to detect objects on different aspect ratios. After that, bounding boxes are visualized on the image with the result matrix. Finally, Non-Maximum Suppression (NMS) operation is applied to remove overlapping bounding boxes.

Table 1: LPYOLO Architecture

| Layer Type | Input | Output | Kernel | Stride | Activation |
|---|---|---|---|---|---|
| QuantConv | 416x416x3 | 416x416x8 | 3x3 | 1 | QuantReLU |
| MaxPooling | 416x416x8 | 208x208x8 | 2x2 | 2 | - |
| QuantConv | 208x208x8 | 208x208x8 | 3x3 | 1 | QuantReLU |
| MaxPooling | 208x208x8 | 104x104x8 | 2x2 | 2 | - |
| QuantConv | 104x104x8 | 104x104x16 | 3x3 | 1 | QuantReLU |
| MaxPooling | 104x104x16 | 52x52x16 | 2x2 | 2 | - |
| QuantConv | 52x52x16 | 52x52x32 | 3x3 | 1 | QuantReLU |
| MaxPooling | 52x52x32 | 26x26x32 | 2x2 | 2 | - |
| QuantConv | 26x26x32 | 26x26x56 | 3x3 | 1 | QuantReLU |
| MaxPooling | 26x26x56 | 13x13x56 | 2x2 | 2 | - |
| QuantConv | 13x13x56 | 13x13x104 | 3x3 | 1 | QuantReLU |
| MaxPooling | 13x13x104 | 13x13x104 | 2x2 | 2 | - |
| QuantConv | 13x13x104 | 13x13x208 | 3x3 | 1 | QuantReLU |
| QuantConv | 13x13x208 | 13x13x56 | 1x1 | 1 | QuantReLU |
| QuantConv | 13x13x56 | 13x13x104 | 3x3 | 1 | QuantReLU |
| QuantConv | 13x13x104 | 13x13x18 | 3x3 | 1 | QuantHardTanh |

## 3.2. Design Flow

The overall design flow is depicted in Figure 1. First of all, the CNN architecture is created with quantized layers and activation functions with various integer bit width precisions. Signed integer and unsigned integer values are used for weights and activations respectively. Also, 8 bits integer values are preferred in the first and last convolution layers in order to get better accuracy. Quantized layers and activation functions are provided by Brevitas library in PyTorch. The created model is trained with quantized values and the trained model is exported in ONNX (Open Neural Network Exchange) format for the intermediate representation. Graph based transformations and optimizations are applied to the ONNX model [16]. In this phase, folding and configuration parameters are set for providing lower latency. How many Processing Element (PE) and Single Instruction Multiple Data (SIMD) will be used on each layer is adjusted by setting folding parameters. Increasing the number of PE and SIMD boosts the parallelism and produces higher speed, but it consumes more logical resources. By setting configuration parameters, the type of memory usage is decided such as BRAM or LUTRAM of the FPGA for each layer. Moreover, the type of logical resource usage is also set such as LUT or DSP of the FPGA for each layer. Afterwards, Intellectual Property (IP) blocks are created for each layer from the ONNX model with Xilinx Vitis HLS tool by using FINN-HLS library which contains layer definitions of the created model in C++. Created IP blocks are stitched together and synthesized via Xilinx Vivado tool. Finally, bitstream and driver are generated to deploy it on PYNQ-Z2 board.

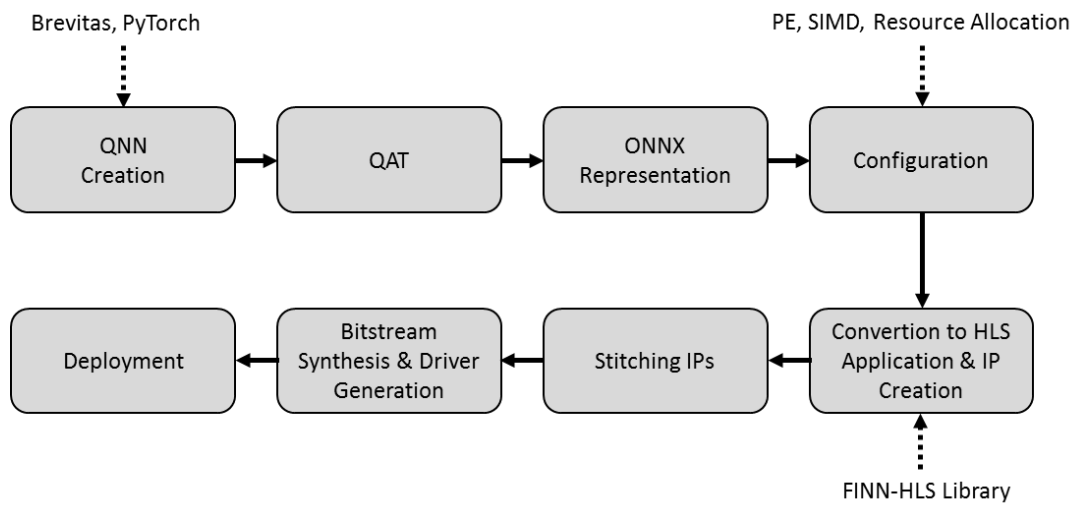

Fig. 1: Design Flow

## 3.3. Inference Flow

Inference operations can be divided into three parts as shown in Figure 2. The first one, preprocessing operations are employed on the PS, Dual Core Arm Cortex A9. The input image or video frame is read, resized to 416x416x3 and packed as UINT8 format. Then, it is given to PL through the Direct Memory Access (DMA) as an input to the created CNN architecture. After the calculations on the PL, the output result matrix is created as 13x13x18 in UINT8 format. This output is taken by the DMA to the PS. At the postprocessing stage, the result matrix is converted to FLOAT32 format by using scale parameters. Then, the matrix is decoded to the bounding boxes of the detected faces and NMS operation is applied. The output image or video frame is generated with the detected faces on it. Finally, the output is streamed via TCP/IP to the host PC or server. Besides, preprocessing, postprocessing and TCP/IP streaming operations are run in different threads in order to increase the throughput.

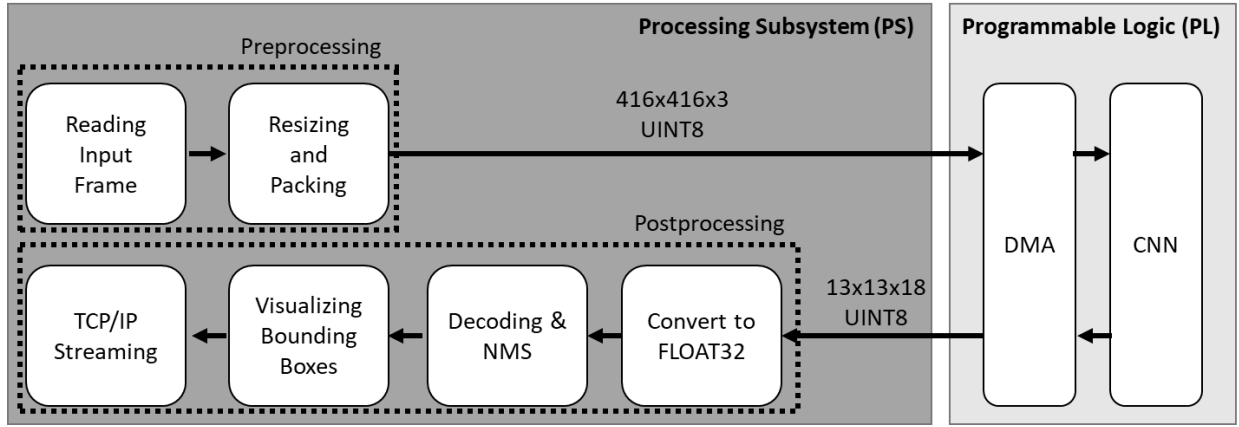

Fig. 2: Inference Flow

## 4. Experimental Results

All the development and measurements are made on TUL Embedded's PYNQ-Z2 board which has Xilinx Zynq 7020 SoC and runs PYNQ version 2.7 (Python Productivity for Zynq) based Ubuntu OS on Dual Core Arm Cortex A9. PL and PS run at frequency of 100 MHz and 667 MHz respectively. Non-Quantized model runs only on the PS and consists of 32 bits floating point values. On the other hand, the quantized models run on PS+PL. Quantized models are created in various bit width precisions such as 2W4A, 3W5A, 4W2A, 4W4A, 6W4A and 8W3A. mWnA means that weights and activations are represented with m and n bit precision respectively.

Table 2: Accuracy Results (mAP)

| Quantization | Easy | Medium | Hard |
|---|---|---|---|
| Non-Quantized | 0,765 | 0,631 | 0,294 |
| 2W4A | 0,671 | 0,469 | 0,199 |
| 3W5A | 0,733 | 0,563 | 0,247 |
| 4W2A | 0,705 | 0,521 | 0,224 |
| 4W4A | 0,757 | 0,590 | 0,261 |
| 6W4A | **0,764** | **0,608** | **0,274** |
| 8W3A | 0,740 | 0,571 | 0,252 |

Accuracy is measured on validation part of WiderFace dataset. The validation dataset is divided into three different categories based on the difficulties: easy, medium and hard images. The dataset contains 32203 images and 393703 faces on them. The dataset organized as follows: 40% for training, 10% for validation and 50% for test images. According to the given results in Table 2, there is a positive correlation between bit precision and accuracy. Also, bit precision of the activations has bigger impact on accuracy. Accuracy difference between Non-Quantized and quantized model is completely acceptable.

Table 3: Latency Results (ms)

| Quantization | Preprocessing | CNN | Postprocessing |
|---|---|---|---|
| Non-Quantized | 18,5 | 5017,9 | 7,5 |
| 2W4A | 35 | 43,5 | 10 |
| 3W5A | 35 | 81,5 | 10 |
| 4W2A | 35 | **37,3** | 10 |
| 4W4A | 35 | 52,3 | 10 |
| 6W4A | 35 | 63,5 | 10 |
| 8W3A | 35 | 49,1 | 10 |

Latency results for a single image are given in Table 3. Quantized models on PS+PL run approximately 50 times faster than the Non-Quantized model on PS. In addition, latency decreases with the lower bit precisions. 4W4A model can be seen as the most suitable model due to having sufficient latency and accuracy results for real-time application on PYNQ-Z2.

Moreover, 18 FPS throughput is achieved on the TCP/IP video streaming with 4W4A quantized model. While video streaming, successive frames are processed in a pipeline structure. Also, this result is gotten with the help of multithreading the operations. Multithreading enables the CPU for processing the operations in different threads. Therefore, the operations run parallel without waiting each other. 2.4 Watt total power consumption on the board is observed. FPGA utilization for the 4W4A model is as follows. 74% of LUT, 11% of LUTRAM, 65% of BRAM, 51% of Flip-Flops, 92% of DSP. Furthermore, the temperature of the SoC is measured as 35.6 °C at runtime.

## 5. Conclusion

In this work, face detection with quantized Yolo model has been successfully implemented on a low-end SoC architecture. Quantization of a model leds to almost 50 times faster operation and tolerable accuracy drop compared to non-quantized model. At the end, very low power consumption, acceptable accuracy rate and sufficient throughput are accomplished for edge computing. Moreover, on-chip memory usage lowered the power surge remarkably and the design is deeply satisfactory for real-time video surveillance applications.

In the future works, MobileNet [31] architecture will be used as backbone because of having depthwise separable convolutions which consume less logical resources compared to standard convolutions. Besides, Neural Architecture Search (NAS) [32] method will be tried for creating an optimized backbone network.